\documentclass[sigconf]{acmart}
\setcopyright{none}
\acmConference[]{}{}{}
\acmBooktitle{}
\acmISBN{}
\renewcommand\footnotetextcopyrightpermission[1]{}
\AtBeginDocument{%
  \providecommand\BibTeX{{%
    Bib\TeX}}}

\acmDOI{}

\usepackage{amsmath,amssymb,amsfonts}
\usepackage{graphicx}
\usepackage{textcomp}
\usepackage{xcolor}
\usepackage{caption}
\usepackage{float}
\usepackage{dblfloatfix}
\usepackage{subcaption}
\usepackage{array}
\usepackage{makecell}
\usepackage{multirow}
\usepackage{tikz}
\usepackage{pgfplots}
\usepackage{pgfplotstable}
\usepackage{subcaption}
\usepackage{soul}
\usepackage{color, colortbl}
\usepackage{url}
\usepackage{ragged2e}
\usepgfplotslibrary{statistics}
\usepackage{graphicx}
\usepackage{standalone}
\usepackage{booktabs}
\usepackage[T1]{fontenc} 
\usepackage{amssymb}
\pgfplotsset{compat=1.18}
\usepackage[ruled,vlined]{algorithm2e}
\usepackage{algpseudocode}
\usepackage{microtype}
\usepackage{enumitem}

\DeclareUnicodeCharacter{2212}{\ensuremath{-}}
\def\BibTeX{{\rm B\kern-.05em{\sc i\kern-.025em b}\kern-.08em
    T\kern-.1667em\lower.7ex\hbox{E}\kern-.125emX}}

\newcommand{\mydiamond}{%
  \sbox0{$\lozenge$}%
  \usebox0\kern-.5\wd0\clap{\raisebox{.1ex}{\scalebox{.7}[1]{$-$}}}\kern.5\wd0%
}

\begin{document}

\title{ConvLUT: Rethinking Convolutional Architectures for Ultra-Low-Latency LUT-Based Inference
}

\author{Marta Andronic}
\email{marta.andronic18@imperial.ac.uk}
\affiliation{%
  \institution{Imperial College London}
  \city{London}
  \country{UK}
}

\author{Olivia Weng}
\email{oweng@ucsd.edu}
\affiliation{%
  \institution{University of California San Diego}
  \city{San Diego}
  \state{CA}
  \country{USA}
}

\author{Danial Zuberi}
\email{dzuberi@ucsd.edu}
\affiliation{%
  \institution{University of California San Diego}
  \city{San Diego}
  \state{CA}
  \country{USA}
}

\author{Ryan Kastner}
\email{kastner@ucsd.edu}
\affiliation{%
  \institution{University of California San Diego}
  \city{San Diego}
  \state{CA}
  \country{USA}
}

\author{George A. Constantinides}
\email{g.constantinides@imperial.ac.uk}
\affiliation{%
  \institution{Imperial College London}
  \city{London}
  \country{UK}
}

\renewcommand{\shortauthors}{Andronic et al.}
\begin{abstract}
Lookup-table (LUT)-based neural networks (NNs) offer a compelling path to ultra-low-latency inference on FPGAs. However, existing LUT-based neural networks have largely been restricted to MLP-style topologies, leaving structured vision and audio tasks underexplored. We present ConvLUT, a family of fully unrolled LUT-mappable convolutional architectures targeting the MLPerf Tiny benchmarks. It is comprised of a standard convolutional model built around a hierarchical tree of LUTs and a depthwise separable model following the MobileNet design philosophy. These architectures are enabled by four key contributions: a unified LUT-based convolutional framework supporting standard, depthwise, and pointwise convolutions under explicit fan-in constraints; a LUT-tree that addresses the $k^2c$ receptive field dimensionality bottleneck; a channel-partitioned LUT pooling layer that replaces conventional spatial pooling; and a hardware-aware iterative pruning strategy for stable convergence at scale. Evaluated on CIFAR-10, Visual Wake Words, and Keyword Spotting, ConvLUT achieves competitive accuracy with fully unrolled inference latencies below $\sim 50$\,ns. On CIFAR-10 it reaches at least a $8844\times$ latency reduction over non-logic-based prior implementations, and outperforms existing MLP-style LUT-based models on the same task by at least $20$\,pp in test accuracy.
\end{abstract}

\maketitle
\pagestyle{plain}

\section{Introduction and Motivation}

LUT-based NNs are a promising alternative to multiply-accumulate-centric architectures, particularly in hardware-constrained environments requiring ultra-low latency. By replacing arithmetic operations with discrete function approximators, LUT-based models eliminate costly multiplications and enable efficient mapping to FPGA platforms.

Differentiable weightless neural networks (DWNs)~\cite{dwn} and differentiable logic gate networks (DiffLogicNets)~\cite{petersen2022deep} train LUT/gate networks using differentiable relaxations of discrete Boolean functions. In contrast, works such as LogicNets~\cite{logicnets} and NeuraLUT~\cite{NeuraLUT} follow a different strategy: neural networks are trained under structural constraints—bounded fan-in and quantized activations—such that the model compiles directly into Logical LUTs (L-LUTs), \textit{i.e.}, lookup tables of arbitrary size rather than FPGA LUT primitives. NeuraLUT-Assemble~\cite{assemble} extends this approach by incorporating tree-structured LUT compositions and the one-shot magnitude pruning from~\cite{poly_extension}, thereby improving scalability in MLP-style architectures.

\begin{figure}[tb]
\centerline{\includegraphics[width=0.85\columnwidth]{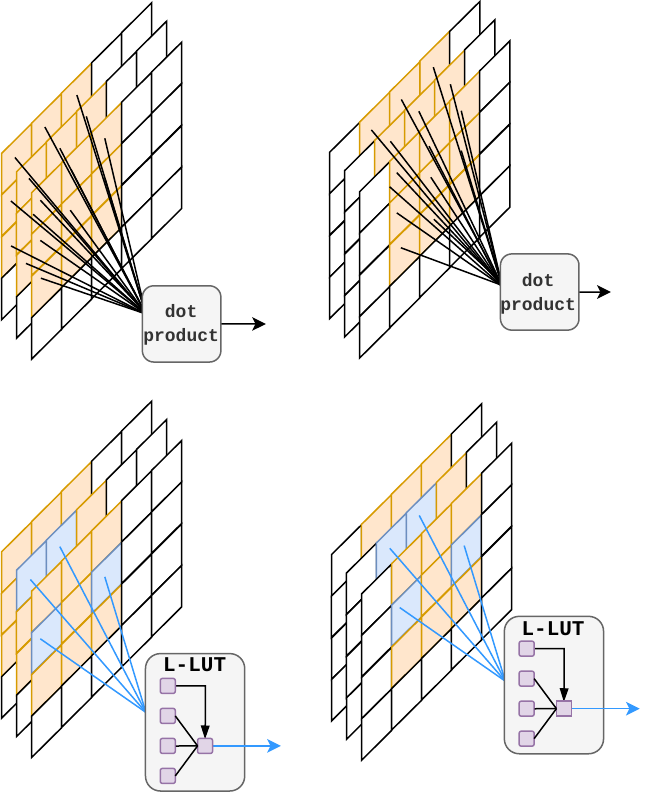}}
\caption{Traditional and proposed convolutional kernels. Top row: a standard dot-product kernel connects every input within the $k^2 c$ receptive field. Bottom row: the proposed L-LUT kernel replaces the dot product with a NeuraLUT surrogate~\cite{NeuraLUT}. Each output channel learns its own sparsity pattern, which is shared across all spatial positions during the sliding operation. Fan-in is kept at $F$ inputs per L-LUT.}
\label{fig:high}
\end{figure}

Despite these advances, existing LUT-based architectures use relatively simple layers and do not scale to tasks that traditionally require convolutional operations~\cite{dwn,assemble}. Logic-gate-based architectures that do support convolutions~\cite{petersen2024convolutional} exist, but their reliance on 2-input gates does not directly exploit the larger fan-in of native FPGA LUT primitives. Thus, extending LUT-based networks to convolutional architectures in a hardware-aware manner remains an open problem.

In NeuraLUT, each L-LUT absorbs a fully connected MLP with skip connections, but is subject to a bounded input fan-in constraint $F$ and bitwidth $B$. Keeping $F$ and $B$ small is crucial for mapping L-LUTs to FPGAs because the L-LUT size grows as $2^{BF}$. Scaling the L-LUT construction to full convolutional networks thus introduces three key challenges. First, a $k \times k$ kernel over $c$ channels induces an input dimensionality of $k^2c \gg F$. If we set $F = k^2c$, we would quickly run out of LUTs on the FPGA. Naïvely pruning to satisfy this constraint severely restricts the receptive field, decreasing accuracy. Second, classical pooling operations feature large fan-in, exploding resource usage. Third, one-shot pruning, sufficient for MLPs, does not converge well at convolutional scale, as seen in our experiments. These three challenges prevent us from constructing convolutional L-LUT models with competitive accuracy within an FPGA's resource constraints.

We address these challenges from an architectural perspective. We develop a unified L-LUT convolutional framework called ConvLUT that supports standard, depthwise, and pointwise convolutions under explicit fan-in constraints, with each output channel maintaining its own learned sparsity pattern shared across spatial positions. In ConvLUT, the standard convolution presents the hardest case: its $k^2c$ input dimensionality far exceeds $F$, requiring a dedicated solution~(Fig.~\ref{fig:high}). We therefore introduce a hierarchical LUT tree construction in which structured sparse pointwise layers compose multiple fan-in-bounded units, aggregating an extended portion of the receptive field while preserving LUT-mappability. Depthwise convolution, whose dimensionality scales only with $k^2$, and pointwise convolution, whose dimensionality scales with $c$, require no such decomposition and can be brought within hardware limits with modest pruning. 

To replace classical pooling, we propose a channel-partitioned LUT-based spatial pooling layer implemented as a structured sparse fully connected layer in which each L-LUT neuron aggregates inputs from a single channel, preserving spatial separation while enabling a richer learned aggregation function.

Scaling these architectures further requires careful sparsification. Direct one-shot pruning, which was sufficient in earlier small-scale NeuraLUT-Assemble models~\cite{assemble}, becomes unusable at larger scales because accuracy falls dramatically. Thus, we introduce a hardware-aware iterative pruning strategy that progressively enforces fan-in constraints during training.

With ConvLUT, we construct two families of LUT-mappable CNN architectures. The first follows a standard convolutional design, leveraging the hierarchical LUT tree. The second adopts a depthwise separable design inspired by MobileNet~\cite{mobilenet}, combining depthwise and pointwise convolutions to reduce computational complexity.

This combination of architectural restructuring and iterative hardware-aware pruning enables NeuraLUT to scale beyond MLP-style networks and tackle MLPerf Tiny benchmarks, including CIFAR-10, Visual Wake Words, and Keyword Spotting. These tasks require spatial or spectro-temporal processing and have historically been difficult to address with LUT-mappable MLPs without significant accuracy degradation. In the spirit of LUT-based neural networks, all models are implemented fully unrolled, instantiating each L-LUT for all spatial windows to achieve ultra-low latency inference. Our focus is to demonstrate that ConvLUT scales to significantly harder tasks than previously attempted.

In this paper, we make the following novel contributions:
\begin{itemize}
    \item{We introduce ConvLUT, a family of LUT-based convolutional architectures, comprising both standard and depthwise separable models, enabled by a unified framework supporting standard, depthwise, and pointwise L-LUT convolutions under explicit fan-in constraints.}
    \item{We propose a hierarchical LUT tree construction for standard convolution, where structured sparse pointwise layers compose fan-in-bounded L-LUTs to extend the effective receptive field without violating hardware constraints.}
    \item{We introduce a channel-partitioned LUT pooling layer that replaces spatial pooling with a learned, channel-wise aggregation.}
    \item{We propose a hardware-aware iterative pruning strategy that progressively enforces fan-in constraints during training, enabling stable convergence at the scale of convolutional networks where one-shot pruning fails.}
    \item{We demonstrate competitive accuracy on three MLPerf Tiny benchmarks---CIFAR-10, Visual Wake Words, and Keyword Spotting---while achieving inference latencies below $\sim$50\,ns, representing a minimum of $1850\times$ reduction over non-LUT-based prior implementations.}
\end{itemize}

\section{Background}
\label{section:background}
\subsection{LUT-Based and Logic-Based Neural Networks}
Ultra-low latency accelerators can be broadly divided into two categories. The first trains models directly in the logic domain using differentiable relaxations of Boolean operators, as in DWN~\cite{dwn} and LUTNet~\cite{wang19lut, wang20}.  DiffLogicNets~\cite{petersen2022deep} constructs networks from primitive operations such as AND and XOR, and have been extended to convolutional architectures~\cite{petersen2024convolutional} through tree-structured compositions of logic gates.

The second category follows a constraint-based approach, as in LogicNets~\cite{logicnets}, PolyLUT~\cite{poly}, PolyLUT-Add~\cite{polyadd}, AmigoLUT~\cite{amigolut}, and NeuraLUT~\cite{NeuraLUT}. Here, neural networks are trained with bounded fan-in and quantized activations, such that the resulting model can be mapped to a circuit of L-LUTs. 

A key distinction between these paradigms lies in how well their computational primitives match the underlying hardware primitives. LUT-based approaches map efficiently to the FPGA fabric, while logic gate networks require synthesis into circuits of larger LUT primitives that can introduce routing overhead at scale. This limitation was observed by the authors of DWN~\cite{dwn}, who reported that neither their larger networks of 2-input gates nor DiffLogicNets could be successfully synthesized. As a result, logic-gate networks may be better suited to ASIC platforms, where such fine-grained primitives are native.

\subsection{Convolutional Neural Networks (CNNs)}
\begin{table}[t]
\centering
\caption{Summary of convolutional notation.}
\label{tab:notation}
\begin{tabular}{cl}
\hline
\textbf{Symbol} & \textbf{Description} \\
\hline
$c$   & Number of input channels \\
$N$   & Number of output channels \\
$k$   & Spatial kernel size ($k \times k$) \\
$h, w$ & Input feature map height and width \\
$F$   & L-LUT input fan-in \\
$B$   & Activation bitwidth \\
\hline
\end{tabular}
\end{table}

CNNs form the foundation of modern vision and audio models. Extending LUT-based neural networks to convolutional architectures presents several fundamental challenges. First, the receptive field of a standard convolution scales as $k^2c$, which is typically much larger than the allowable fan-in $F$. Enforcing this constraint through pruning severely restricts each neuron's connectivity, limiting its ability to capture information from the full receptive field.

The PolyLUT~\cite{poly_extension} extension introduced a hardware-aware regularization scheme that incorporates fan-in constraints into training and performs one-shot pruning. NeuraLUT-Assemble~\cite{assemble} extends the scalability of this framework by introducing tree-structured compositions of L-LUTs, enabling the effective fan-in of a neuron to be increased without exceeding hardware limits.
Unfortunately, neither approach generalizes directly to convolutional layers, where the receptive field is shared across spatial locations and channels.

Second, classical pooling operations are incompatible with LUT-based implementations because they require aggregating over large input windows, leading to fan-in requirements that exceed what can be implemented efficiently using LUTs.

Taken together, these limitations highlight a gap in existing approaches. Methods that enforce hardware constraints, such as NeuraLUT or NeuraLUT-Assemble~\cite{assemble}, do not scale to convolutional architectures, and methods that support convolutional structures, such as DiffLogicNets~\cite{petersen2024convolutional}, do not explicitly align with FPGA LUT primitives. 

\subsection{Hardware Constraints for LUT-based Inference}
LUT-based inference imposes strict architectural constraints that fundamentally shape network design. Each neuron is implemented as an L-LUT with fan-in $F$ and input bitwidth $B$ making the corresponding truth table contain $2^{BF}$ entries. This exponential scaling makes large fan-in infeasible in practice, requiring networks to operate under tight connectivity budgets. Consequently, the primary design challenge is not computational complexity within each LUT but the ability to express the desired function under strict fan-in and connectivity constraints while maintaining efficient mapping to hardware. 

In addition, LUT-based designs are typically deployed in a fully unrolled configuration, where all operations are instantiated spatially and executed in parallel. In this regime, latency is determined by circuit depth rather than sequential computation, and throughput is maximized by eliminating control flow and data reuse.

\begin{figure*}[tb]
\centerline{\includegraphics[width=0.8\textwidth]{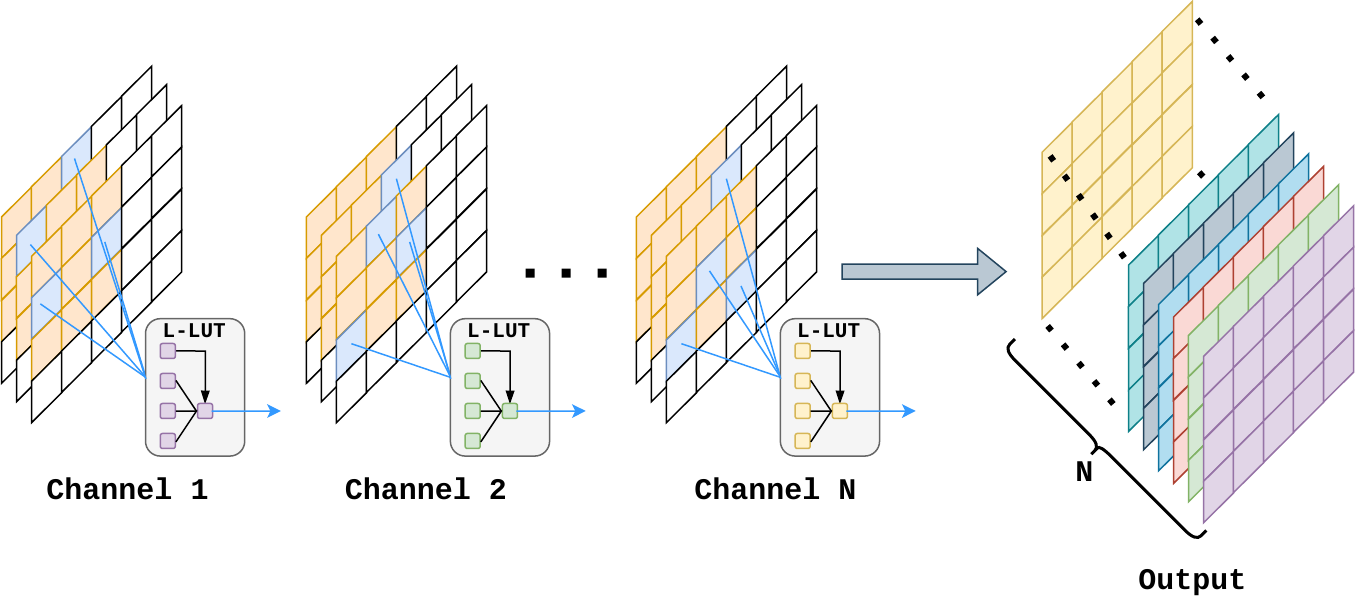}}
\caption{A sparse L-LUT convolutional layer with $3$ input channels and $N$ output channels. Each channel maintains its own learned sparsity pattern over the $k^2 c$ receptive field. The resulting output volume has $N$ channels, each encoding a distinct sparse projection of the local neighbourhood.}
\label{fig:channels}
\end{figure*}

\begin{figure*}
\centering
\hfill
\begin{subfigure}{0.18\textwidth}
    \includegraphics[width=\textwidth]{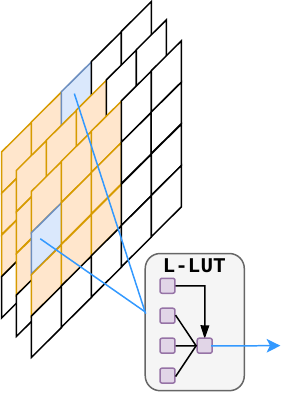}
    \caption{Standard conv.}
    \label{fig:first}
\end{subfigure}
\hfill
\begin{subfigure}{0.18\textwidth}
    \includegraphics[width=\textwidth]{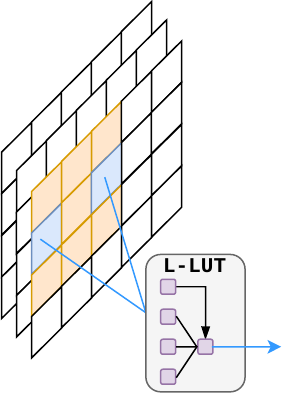}
    \caption{Depthwise conv.}
    \label{fig:second}
\end{subfigure}
\hfill
\begin{subfigure}{0.18\textwidth}
    \includegraphics[width=\textwidth]{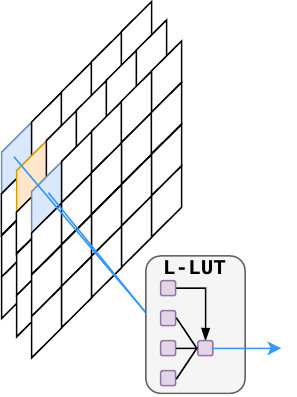}
    \caption{Pointwise conv.}
    \label{fig:third}
\end{subfigure}
\hfill
\caption{Convolution types supported by ConvLUT. 
In standard convolution~(\subref{fig:first}), each L-LUT neuron selects a sparse subset of the full $k^2 c$ receptive field. In depthwise convolution~(\subref{fig:second}), neurons are restricted to a single input channel, with sparsity over the $k^2$ spatial locations. In pointwise convolution~(\subref{fig:third}), neurons operate across all channels at a single spatial location, with sparsity over the channel dimension. In all cases, each output channel maintains its own learned sparsity pattern, which is shared across spatial positions during the sliding operation.}
\label{fig:architectures}
\end{figure*}

\section{ConvLUT}
\subsection{L-LUT Convolution}

A standard convolutional layer computes each output activation as a dot product between a learned weight vector and a local input patch. In ConvLUT, we replace this dot product with a NeuraLUT computational block~\cite{NeuraLUT}, which is trained using a skip-connected MLP as a differentiable surrogate and compiled into an L-LUT for inference (shown in Fig.~\ref{fig:high}). 

A key insight of this construction is that the sparsity pattern and the L-LUT function can be shared across spatial positions during the sliding operation, directly analogous to weight sharing in standard convolutions. As in Fig.~\ref{fig:channels}, each of the $N$ output channels has its own distinct learned sparsity pattern and function over the $k^2c$ receptive field, but applies it identically at every spatial location. In a fully unrolled implementation, one L-LUT is instantiated per output channel per spatial window, achieving throughput equal to clock frequency. This unrolling factor, equal to the number of spatial windows, directly determines the resource cost of each layer~(Table~\ref{tab:architectures}).

We extend this construction to all three convolution types used in modern efficient architectures, illustrated in Fig.~\ref{fig:architectures}. Standard convolution assigns each L-LUT neuron a sparse subset of the full $k^2c$ receptive field, spanning both spatial and channel dimensions. Depthwise convolution restricts each neuron to a single input channel, with sparsity over the $k^2$ spatial locations only, naturally reducing receptive field dimensionality. Pointwise convolution operates across all input channels at a single spatial location, with sparsity over the channel dimension. Supporting all three types within a unified fan-in-constrained framework is a prerequisite for realizing CNN architectures as fully LUT-mappable networks.

Although our analysis and results use NeuraLUT as the LUT primitive, the architectural principles introduced in this work are not specific to this formulation. Any LUT-based surrogate can be embedded within ConvLUT. Our contributions therefore extend beyond NeuraLUT and provide general guidelines for scaling LUT-based models to structured vision and audio tasks.
\subsection{LUT Tree Convolution}
\label{sec:lut_tree}

A key challenge in LUT-based convolution is the receptive field dimensionality. In standard convolution, each output neuron receives $k^2 c$ inputs. Since each neuron is implemented as an L-LUT with a fixed fan-in $F \ll k^2 c$, aggressive pruning is required to reduce connectivity to within the hardware budget. When $k^2 c$ is large, the information discarded by pruning is substantial, limiting the expressiveness of each layer.

\begin{figure*}[tb]
\centerline{\includegraphics[width=\textwidth]{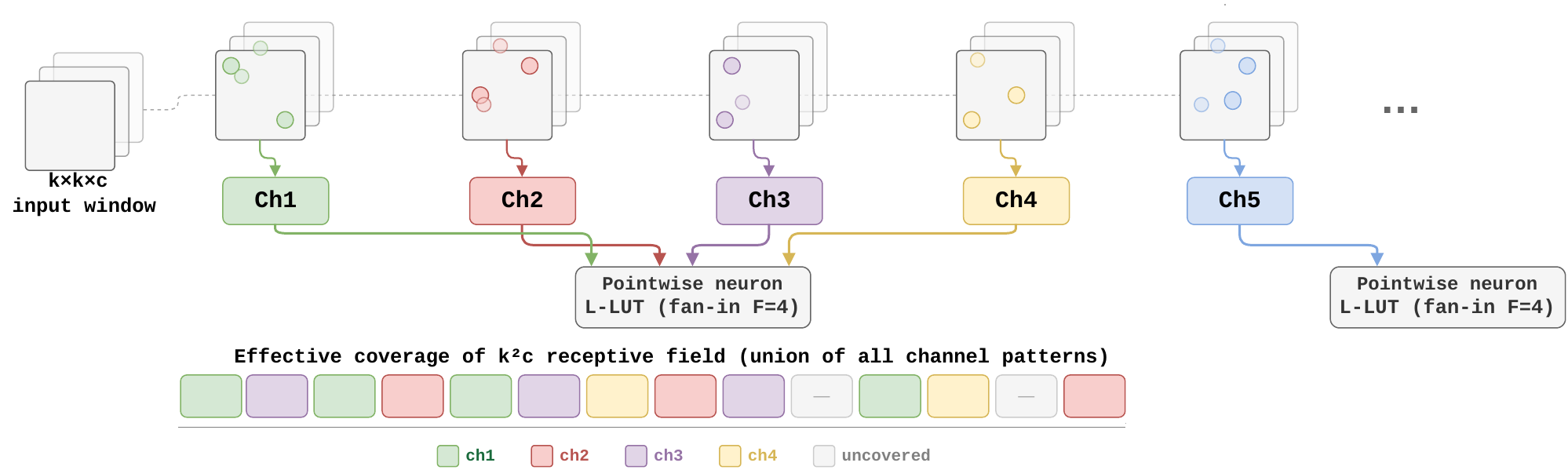}}
\caption{\textbf{LUT Tree Convolution:} A standard convolutional layer produces $N$ output channels, each learning a distinct sparse pattern over the shared $k^2 c$-dimensional receptive field (colored dots indicate active connections per channel). A subsequent pointwise layer with fixed structured sparsity aggregates groups of $G = F$ channels into single L-LUT neurons (each channel occupies one fan-in input, so the group size is bounded by the fan-in budget $F$). The effective receptive field (bottom bar) covers the union of all those channel patterns and some locations remain uncovered due to the finite fan-in budget $F$.}
\label{fig:tree}
\end{figure*}

To address this, we propose \textit{LUT tree convolution}: a multi-layer construction that expands the effective receptive field without violating the fan-in constraint. The key observation is that in a standard convolutional layer, all output channels share the same spatial receptive field. Multiple output channels can therefore learn \textit{different} sparse connectivity patterns over the same $k^2 c$-dimensional input patch, each capturing a complementary view of the local neighborhood. A subsequent pointwise layer with structured fixed sparsity aggregates groups of these channels into each output neuron, forming the second level of the tree (Fig.~\ref{fig:tree}). Since each input channel to the pointwise layer has already distilled a distinct sparse projection of the shared receptive field, the pointwise neuron effectively combines information from a much larger portion of the original patch than any single L-LUT could handle.

Formally, let output channel $n$ have sparsity mask $\mathbf{m}_n \in \{0,1\}^{k^2 c}$ with $F$ non-zero entries. The pointwise layer groups channels into blocks of size $G$, with each output neuron connected to one block via a fixed structured mask. Since the L-LUT is constrained by fan-in $F$, at most $F$ channels can be aggregated, \textit{i.e.}\ $G \leq F$. We therefore set $G = F$. The effective receptive field therefore spans up to $G \cdot F = F^2$ input locations, compared to $F$ for a single-stage L-LUT (Fig.~\ref{fig:tree}). This construction can be extended by stacking additional pointwise layers to further expand the receptive field.

This construction is specific to standard convolutions. In depthwise convolutions, each output channel operates over a single input channel, so different output channels correspond to entirely different input patches. 
Aggregating across depthwise channels via a pointwise layer does not produce the same receptive field expansion, as the channels carry spatially disjoint information.

\subsection{LUT-Based Spatial Pooling Layer}
\label{sec:lut_pooling}
 \begin{figure*}[tb]
\centerline{\includegraphics[width=0.9\textwidth]{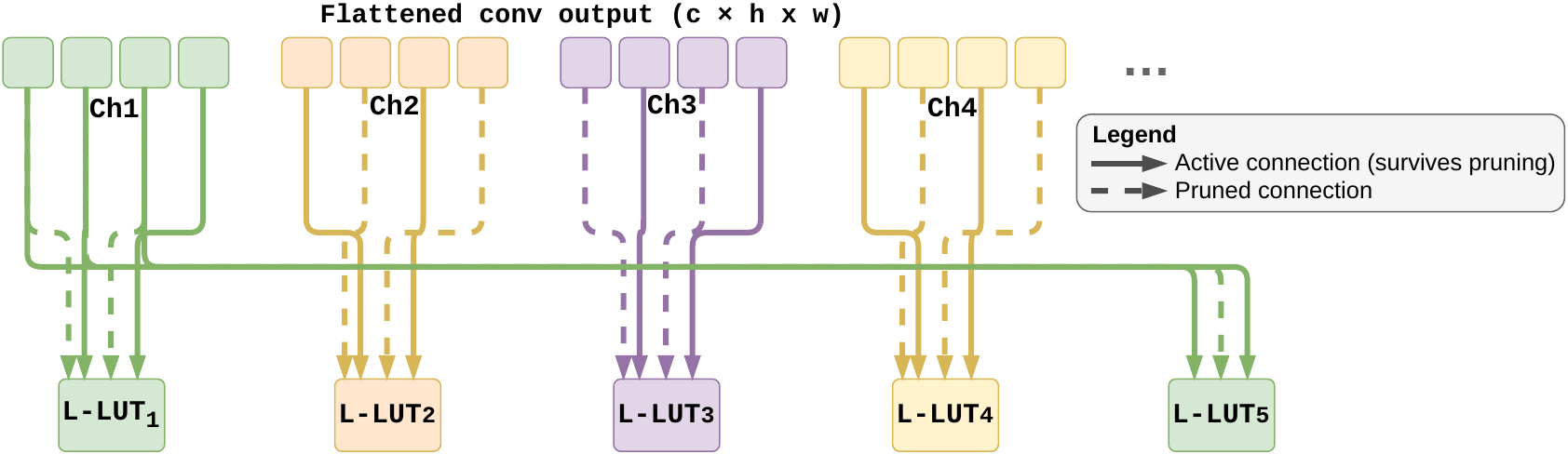}}
\caption{\textbf{Channel-partitioned LUT pooling layer:} The flattened convolutional output is partitioned by channel, with each L-LUT neuron constrained to receive inputs exclusively from a single channel's spatial feature map. Output neurons are assigned to channels in an interleaved fashion, so that multiple neurons attend to each channel, compensating for information lost through pruning. Solid arrows indicate active connections after pruning to fan-in $F$; dashed arrows indicate pruned connections. Each neuron implements a learned sparse spatial aggregation via the NeuraLUT surrogate~\cite{NeuraLUT}, replacing fixed max or average pooling.}
\label{fig:pooling}
\end{figure*}
 
In LUT-based frameworks, deploying a conventional pooling layer presents a fundamental incompatibility: average and max pooling operate over potentially large input windows that do not map naturally onto the L-LUT abstraction, requiring fan-ins far beyond the 4-6 inputs practical for FPGA implementation. A direct implementation would introduce a bottleneck that undermines the uniform, ultra-low latency design that makes LUT-based inference attractive in the first place.

We propose replacing the pooling layer with a structured, sparse fully connected L-LUT layer (Fig.~\ref{fig:pooling}). This approach offers two complementary advantages. First, by constraining each output neuron to receive inputs exclusively from the spatial feature map of a single convolutional channel, we preserve the channel-wise spatial aggregation behavior of classical pooling. Second, rather than applying a fixed max or average operation, each neuron implements a learned surrogate function, enabling increased expressivity in spatial aggregation. The result is a pooling replacement that is both architecturally consistent with the surrounding LUT layers and capable of learning richer aggregation functions than standard fixed pooling operators.

 \begin{figure}[tb]
\centerline{\includegraphics[width=\columnwidth]{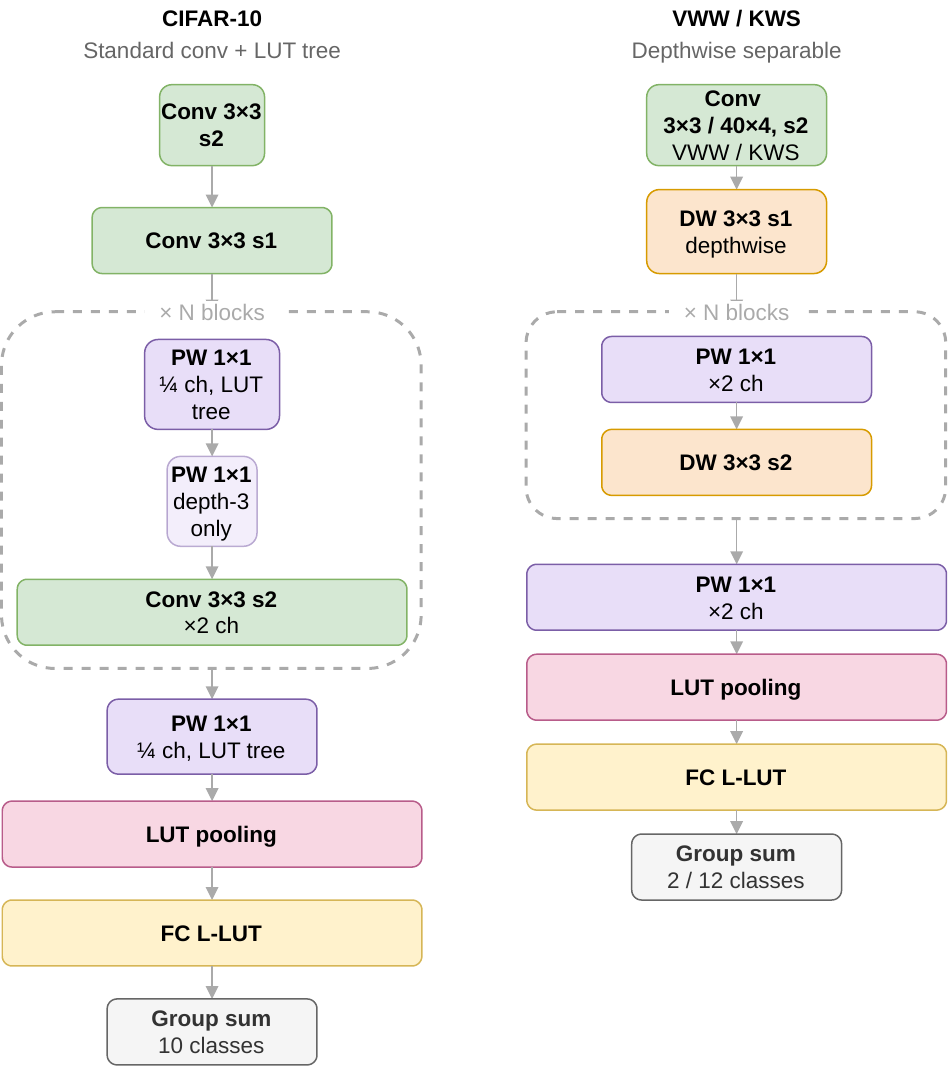}}
\caption{Architecture overview of the two model families. Left: the CIFAR-10 CNN model, in which each block comprises one or two pointwise layers acting as LUT tree aggregators, as seen in Section~\ref{sec:lut_tree}, followed by a strided standard convolution. The faded second pointwise layer is present only in the depth-3 variant. Right: the DS-CNN model used for VWW and KWS, following the MobileNet~\cite{mobilenet} structure, in which each block pairs a pointwise layer for cross-channel mixing with a strided depthwise convolution for spatial downsampling. The LUT tree construction does not apply here. Both architectures share the same classifier head: a channel-partitioned LUT pooling layer (as described in Section~\ref{sec:lut_pooling}), a fully connected L-LUT layer, and a group sum producing the final class scores.}
\label{fig:arch}
\end{figure}

\subsubsection{Channel-Partitioned Connectivity Mask}
 
Let the input after the final convolutional layer have $c$ channels and a spatial resolution of $h \times w$, yielding a flattened feature vector of dimension $c \cdot h \cdot w$. The subsequent fully connected layer maps this to $N$ output neurons. We define a binary connectivity mask $\mathbf{M} \in \{0,1\}^{N \times c \cdot h \cdot w}$ that enforces a strict channel-partitioned structure. Specifically, we partition the $N$ output neurons into $c$ interleaved groups of $K = N / c$ neurons each, enforcing that c divides N. Neuron $i$ is assigned to channel $c_i = i \bmod c$, and its receptive field is restricted to the $P = h \cdot w$ inputs corresponding to that channel:
 
\begin{equation}
    \mathbf{M}_{i,j} = \begin{cases} 1 & \text{if } \lfloor j / P \rfloor = (i \bmod c) \\ 0 & \text{otherwise} \end{cases}
\end{equation}
 
where $j \in \{0, \ldots, c \cdot P - 1\}$ indexes the flattened input. This interleaved assignment ensures that each channel is attended to by exactly $K$ output neurons, distributing representational capacity across channels.
 
\subsubsection{Learned Sparse Spatial Aggregation}
 
Within each neuron's receptive field of $P$ spatial inputs, the connectivity is further sparsified through our iterative magnitude pruning. Starting from the full channel receptive field, connections are pruned to a target fan-in $F \ll P$.
 
The redundancy introduced by assigning $K > 1$ neurons to each channel is intentional. Since each neuron ultimately observes only $F$ of the $P$ spatial locations, multiple neurons per channel allow the network to collectively cover the spatial extent of each feature map more richly. Different neurons assigned to the same channel learn to specialize on different sparse subsets of spatial locations, providing complementary views of each channel's feature map.

\subsection{Architectures}

\subsubsection{CIFAR-10: Standard Convolutional Models}
 
The CIFAR-10 architecture follows a plain convolutional backbone in which all spatial mixing is performed by standard $3\times3$ L-LUT convolutional layers, interleaved with pointwise layers that serve as LUT tree aggregators as described in Section~\ref{sec:lut_tree}. The network takes a $32\times32$ RGB image as input and progressively reduces the spatial resolution through strided convolutions while expanding the channel dimension (Fig.~\ref{fig:architectures}). 

The backbone begins with two standard $3\times3$ convolutional layers. The first applies a stride of two to halve the spatial resolution; the second operates at stride one. The remainder of the backbone consists of
repeated blocks, each comprising one or more $1\times1$ pointwise layers followed by a strided $3\times3$ standard convolution. The pointwise layers act as the aggregation stage of the LUT tree, combining the
complementary sparse views of the receptive field assembled by the preceding standard convolution. The strided convolution then halves the spatial resolution and doubles the number of channels, following the standard practice of trading spatial resolution for channel depth as the network deepens. A final pointwise layer projects the features to the desired channel count before the spatial dimensions are collapsed.

We consider two variants that differ in the depth of the LUT tree~(Fig.~\ref{fig:arch}). In the first variant, each block contains a single pointwise layer, forming a depth-2 tree. In the second variant, each block contains two consecutive pointwise layers, forming a depth-3 tree, allowing a larger effective receptive field.
 
After the convolutional backbone the feature map is flattened and passed through the channel-partitioned LUT pooling layer described in Section~\ref{sec:lut_pooling}, which replaces the conventional spatial pooling operation. A fully connected L-LUT layer follows, and the final classification scores are produced by a group sum over the output neurons, with one group per class.
 
\subsubsection{VWW and KWS: Depthwise Separable Models}
 
The depthwise separable models (Fig.~\ref{fig:arch}) take inspiration from MobileNet~\cite{mobilenet}, factorizing each standard convolution into a depthwise convolution that performs spatial mixing within each channel independently, followed by a pointwise convolution that mixes across channels at a single spatial location. This factorization reduces the receptive field dimensionality from $k^2 c$ to $k^2 + c$,  substantially reducing the information discarded by fan-in pruning and making the architecture well suited to highly resource-constrained LUT budgets. Note that because depthwise convolutions operate channel-wise, the LUT tree construction of Section~\ref{sec:lut_tree} does not apply here.
 
The backbone opens with a standard convolutional layer, followed by a depthwise convolution at stride one. The remainder of the backbone consists of repeated depthwise-separable blocks, each comprising a pointwise layer followed by a strided depthwise convolution. Each strided convolution halves the spatial resolution and each pointwise convolution doubles the channel count, progressively concentrating spatial information into the channel dimension. A final pointwise layer produces the output feature map before the spatial dimensions are collapsed.
 
The VWW and KWS models differ primarily in their input modality and first-layer design. The VWW model takes $96\times96$ RGB images as input, and the initial convolution uses a standard $3\times3$ kernel. The KWS model operates on single-channel mel spectrograms and uses a $40\times4$ first-layer kernel, whose tall, narrow shape is tailored to the frequency--time structure of the spectrogram representation. 
 
Following the convolutional backbone, both models pass the flattened features through the channel-partitioned LUT pooling layer, a fully connected L-LUT layer, and a final group sum.

\subsection{Iterative pruning}
\label{sec:pruning}

\begin{algorithm}[t]
\caption{Hardware-aware iterative pruning. The schedule $\mathcal{S}$ progressively tightens the fan-in across $T$ stages, $F_1 > F_2 > \cdots > F_T = F^*$, to the target fan-in $F^*$. After each prune step, the model retrains for $E_t$ epochs with masks frozen, allowing weights to recover before connectivity is reduced further.}
\label{alg:iterative_pruning}

\KwIn{Model $\mathcal{M}$; pruning schedule
$\mathcal{S} = \{(F_t,\, E_t)\}_{t=1}^{T}$,
where $F_t$ is the fan-in target and $E_t$ the number of
retraining epochs for stage $t$, with $F_1 > \cdots > F_T = F^*$}

\KwOut{All masks $\mathbf{M}^{(l)}$ satisfy the hardware fan-in $F$}

\ForEach{stage $(F_t, E_t) \in \mathcal{S}$}{
    \ForEach{prunable layer $l$, each neuron $n$}{
        $\mathbf{M}^{(l)}_n \gets
        \operatorname{TopK}\!\left(
        \left|\mathbf{w}^{(l)}_n \odot \mathbf{M}^{(l)}_n\right|,\, F_t
        \right)$\;
    }
    \For{$\text{epoch} = 1$ \KwTo $E_t$}{
        Train $\mathcal{M}$ with $\mathcal{L}_{\mathrm{CE}} +
        \,\Omega(W)$, masks frozen\;
    }
}

Fine-tune $\mathcal{M}$ without regularization ($\,\Omega(W)$)\;

\end{algorithm}

Direct one-shot pruning, sufficient for the smaller MLP-style models in NeuraLUT-Assemble~\cite{assemble}, becomes unstable at convolutional scale. We therefore adopt a hardware-aware iterative pruning strategy, summarized in Algorithm~\ref{alg:iterative_pruning}.

Rather than enforcing the target fan-in $F^*$ immediately, the pruning schedule progressively tightens connectivity across $T$ stages. At each stage, pruning is applied to all prunable layers, \textit{i.e.}, all layers except the pointwise layers part of the LUT tree construction. For each neuron $n$ in layer $l$, only the $F_t$ highest-magnitude active connections are retained, and the mask is updated based on the currently active weights $\mathbf{w}^{(l)}_n \odot \mathbf{M}^{(l)}_n$.

Following each prune step, the model retrains for $E_t$ epochs with all masks frozen, minimizing the objective
\begin{equation}
    \mathcal{L} = \mathcal{L}_{\mathrm{CE}} + \Omega(W),
    \label{eq:loss}
\end{equation}
where $\mathcal{L}_{\mathrm{CE}}$ is cross entropy loss and $\Omega(W)$ is the exponential regularizer of PolyLUT~\cite{poly_extension}. 

After the final stage reaches $F^*$, the model undergoes a final fine-tune without regularization.

\section{Toolflow}
ConvLUT builds upon the NeuraLUT toolflow~\cite{NeuraLUT}, which provides an end-to-end pipeline from quantization-aware training (QAT) through L-LUT conversion, and RTL generation and compilation. The implementation has been extended to support convolutional layers, the channel-partitioned pooling layer, and the iterative pruning described in Section~\ref{sec:pruning}.

\subsubsection{QAT}
Training is implemented in PyTorch. The model is trained using the Adam optimizer with a cosine annealing schedule, with quantization applied via Brevitas~\cite{brevitas}. Input bitwidth $B_{\mathrm{in}}$, hidden bitwidth $B_h$, output bitwidth $B_{\mathrm{out}}$, and fan-in $F$ are specified per model as listed in Table~\ref{tab:architectures}.

\subsubsection{L-LUT conversion and verification}
After training, each L-LUT is materialized by exhaustive enumeration: all $2^{B \cdot F}$ input combinations are evaluated through the trained sub-network to populate the lookup table. Functional correctness is verified at two levels: first, by running inference through the LUT logic model in place of PyTorch operations and confirming the accuracy; second, by using an Icarus Verilog~\cite{iverilog} simulation of the generated Verilog to validate the RTL against the same test vectors.

\subsubsection{RTL generation and synthesis}
Synthesis and place-and-route are performed using Vivado~2020.1, with \texttt{PerformanceOptimized} and \texttt{out-of-context}. This follows prior works ~\cite{logicnets,poly,NeuraLUT,amigolut,treelut}. The goal is to isolate the neural network computation from platform-dependent infrastructure, such as memory controllers and external I/O, enabling consistent comparisons of the computational designs. Consequently, reported latency characterizes the implemented inference datapath rather than complete system-level inference latency.

\section{Evaluation metrics}
\label{section:results}

ConvLUT adopts the established open-source hardware implementation and evaluation framework used by prior ultra-low-latency LUT-based neural networks\cite{logicnets, poly, polyadd, assemble, amigolut}, including register placement, out-of-context (OOC) synthesis, latency and throughput definitions, and FPGA resource reporting. This deliberate choice preserves direct comparability with prior work while isolating the impact of ConvLUT's architectural and training innovations.

Our evaluation addresses a fundamental question: can the ultra-low-latency LUT-based inference paradigm scale beyond relatively simple, MLP-oriented architectures to substantially more complex CNN workloads while preserving its defining hardware properties? Specifically, we investigate whether LUT-based inference can support MLPerf Tiny~\cite{mlperftiny} workloads while maintaining nanosecond-scale latency, deterministic execution, and fully-unrolled computation.

The objective is therefore not to introduce a new hardware evaluation methodology, but to quantify the architectural implications of extending the established LUT-based paradigm to convolutional networks. In particular, our evaluation demonstrates the latency and area trade-offs associated with replacing conventional MAC-based convolution with LUT-based computation.

ConvLUT employs a fully-unrolled architecture, with computational operators spatially instantiated rather than time-multiplexed. With the adopted register placement and pipelining, this enables an initiation interval of one clock cycle (II=1), allowing a new input to be accepted every clock cycle. We distinguish inference latency, defined by the number of pipeline stages and clock period, from throughput, determined by the initiation interval and clock frequency.

We additionally report Area×Delay Product (ADP), following prior LUT-based work, as a combined measure of the two principal optimization objectives: FPGA resource utilization and inference latency. Since fully-unrolled implementations inherently trade increased spatial resources for reduced latency, ADP provides a complementary metric for characterizing this trade-off alongside the individual area and latency measurements.

\section{Evaluation}
\label{section:results}

We evaluate ConvLUT on 3 MLPerf Tiny benchmarks~\cite{mlperftiny}:\smallskip

\noindent\textbf{CIFAR-10}~\cite{cifar10}: $32\times32$ RGB images, 10 classes. The 50,000 training images are split 80/20 into train and validation subsets; the 10,000-image test set is held out for final evaluation.

\smallskip
\noindent\textbf{Keyword Spotting (KWS)}: The MLPerf Tiny adaptation of Google Speech Commands~\cite{speechcommands}, accessed via 
TensorFlow Datasets~\cite{tfds_speechcommands}, targeting 12 classes. The training set is class-imbalanced; we apply loss scaling to compensate. Audio is preprocessed into mel spectrograms.

\smallskip
\noindent\textbf{Visual Wake Words (VWW)}: 
\texttt{vw\_coco2014\_96}~\cite{chowdhery2019visual}, distributed as part of the MLPerf Tiny benchmark. Binary person detection at $96\times96$ resolution, split 80/10/10 into train, validation, and test subsets.

\subsection{Comparison with prior work}
We evaluate ConvLUT against two complementary classes of prior work: ultra-low-latency LUT-based neural networks and conventional FPGA CNN accelerators. The former establish the implementation methodology and hardware regime that ConvLUT seeks to preserve, while the latter represent the more complex application domain that ConvLUT aims to address. No single baseline fully captures both dimensions.

Accordingly, comparisons with prior LUT-based networks demonstrate the architectural progression toward more complex workloads, whereas comparisons with conventional CNN accelerators quantify the latency and resource trade-offs of adopting a fully-unrolled LUT-based implementation.

These comparisons are intended to characterize distinct architectural design points rather than imply equivalent implementation constraints. In particular, ConvLUT prioritizes ultra-low inference latency through extensive spatial parallelism, at the expense of increased hardware resource utilization.

Across all benchmarks, ConvLUT achieves the lowest reported area–delay product, reaching latencies as low as $9$\,ns (Tabs.~\ref{tab:cifar}–\ref{tab:kws}–\ref{tab:vww}). Model architecture details are provided in Tab.~\ref{tab:architectures}. For each model, we report two operating points: a minimally pipelined design minimizing latency, and a more pipelined variant trading latency for higher frequency. Here, latency is reported as pipeline depth times clock period, while throughput assumes an initiation interval of one cycle (II=1).

On CIFAR-10 (Tab.~\ref{tab:cifar}), ConvLUT achieves up to 80.08\% accuracy, and latency as low as $9$\,ns across different operating points. Compared to the fastest arithmetic-based FPGA accelerator (FINN~\cite{finn}), ConvLUT reduces latency by at least $\sim$$8844\times$ at comparable accuracy. Compared to LUT-based neural networks such as DWN~\cite{dwn}, ConvLUT improves accuracy by over 20 percentage points while maintaining nanosecond-scale latency.

\begin{figure}[tb]
\centering
\begin{subfigure}[t]{0.5\textwidth}
    \centering
    \includegraphics[width=\linewidth]{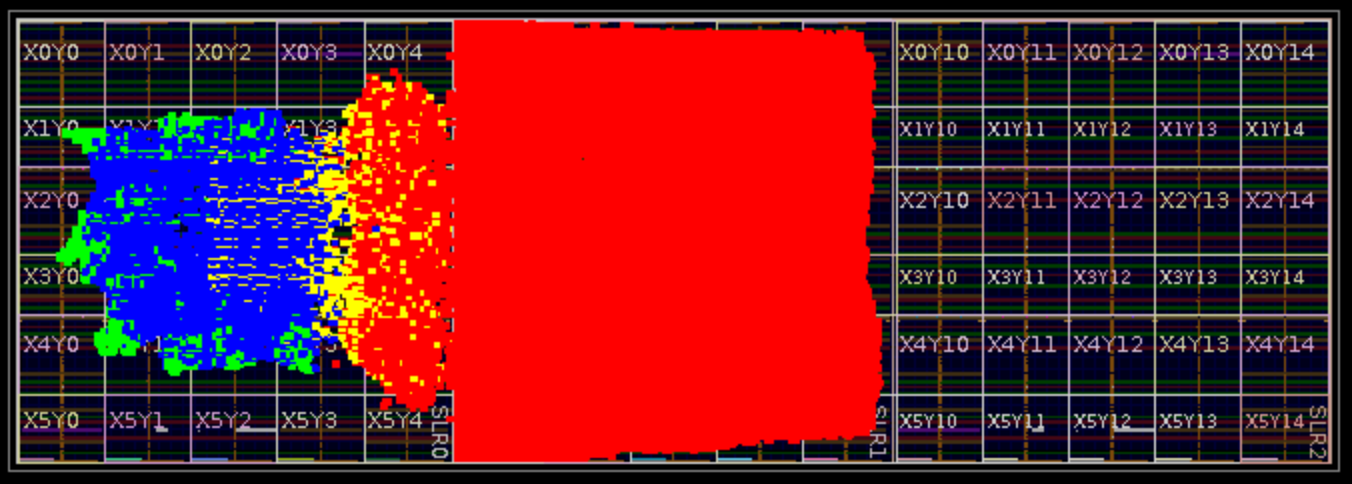}
    \caption{FPGA floorplan of the CNN-S model on a Xilinx VU9P. Convolutional layers are in red, the LUT pooling in yellow, the fully connected layer in blue, and group sum in green.}
\end{subfigure}\hfill%
\begin{subfigure}[t]{0.5\textwidth}
\centering
\begin{tikzpicture}
\begin{axis}[
    ybar stacked,
    bar width=14pt, 
    width=\columnwidth, 
    height=4.1cm, 
    enlarge x limits=0.08,
    ylabel={LUT count},
    ylabel style={font=\footnotesize},
    xtick=data,
    xticklabels={
        L0,L1,L2,L3,L4,L5,L6,L7,L8,L9
    },
    xticklabel style={font=\footnotesize, anchor=north},
    yticklabel style={font=\footnotesize},
    scaled y ticks=false,
    ytick={0,20000,40000,60000,80000,100000},
    yticklabels={0, 20k, 40k, 60k, 80k, 100k},
    ymin=0, ymax=110000,
    axis lines*=left,
    tick align=outside,
    ymajorgrids=true,
    grid style={dotted, gray!40},
    legend style={
        at={(0.98,0.98)},
        anchor=north east,
        font=\footnotesize,
        draw=none,
        fill=none,
        legend columns=1
    },
    legend cell align=left,
]
\addplot[fill=red, draw=none]
coordinates {
    (0, 23816) (1, 99567) (2, 43396) (3, 49923) (4, 14237)
    (5, 22019) (6, 6514) (7, 0) (8, 0) (9, 0)
};
\addlegendentry{Conv layers}

\addplot[fill=yellow, draw=none]
coordinates {
    (0, 0) (1, 0) (2, 0) (3, 0) (4, 0)
    (5, 0) (6, 0) (7, 13920) (8, 0) (9, 0)
};
\addlegendentry{LUT Pooling Layer}

\addplot[fill=blue, draw=none]
coordinates {
    (0, 0) (1, 0) (2, 0) (3, 0) (4, 0)
    (5, 0) (6, 0) (7, 0) (8, 26731) (9, 0)
};
\addlegendentry{FC Layer}

\addplot[fill=green, draw=none]
coordinates {
    (0, 0) (1, 0) (2, 0) (3, 0) (4, 0)
    (5, 0) (6, 0) (7, 0) (8, 0) (9, 8818)
};
\addlegendentry{Group Sum}
\end{axis}
\end{tikzpicture}
\caption{Per-layer LUT utilization for the CNN-S model on a Xilinx VU9P.}
\label{fig:lut_utilisation}
\end{subfigure}

\caption{Implementation results for the CNN-S model. By designing the network architecture around the native LUT primitive, the resulting circuit maps compactly and efficiently onto the FPGA fabric, demonstrating that co-designing neural network architectures with the target hardware abstraction is key to achieving high resource efficiency and ultra-low latency.}
\label{fig:cnn_s_impl}
\end{figure}

Fig.~\ref{fig:cnn_s_impl} shows that the design naturally maps compactly onto the FPGA fabric, showing the advantage of closely aligning the model structure with the underlying LUT primitives. This alignment directly enables the achieved timing and allows successful placement and routing even at extreme utilization. In particular, we are able to place and route the largest CIFAR-10 model despite exceeding 95\% LUT utilization.

On KWS (Tab.~\ref{tab:kws}), ConvLUT achieves state-of-the-art 91.53\% accuracy while having the lowest latencies and area-delay products. Compared to LUT-based neural networks such as DWN, ConvLUT again improves accuracy by over 20 percentage points. Compared to prior FPGA accelerators, such as fpgaconvnet~\cite{fpgaconvnet_results}, ConvLUT provides comparable accuracy while reducing latency by over 4 orders of magnitude.

On VWW (Tab.~\ref{tab:vww}), ConvLUT achieves the MLPerf Tiny accuracy target of $80\%$, which is lower than fpgaconvnet by $4.8$\,pp. Latency is reduced by over $10^6\times$ and area–delay product by $5$ orders of magnitude.

Although LUT utilization is high, this reflects the fully unrolled design required for ultra-low latency. The resulting increase in parallelism comes at the cost of area, but the LUT-based design eliminates the need for DSPs and BRAM, with the entire computation mapped to LUT fabric. While this increases area, the latency gains more than offset the cost, yielding the lowest area–delay products across all benchmarks.

\begin{table*}[t]
\centering
\small
\caption{Architecture configurations. The unrolling factor denotes the number of spatial windows per resolution stage, i.e.\ the number of parallel L-LUT instances implementing the same sparse kernel. The total number of L-LUTs equals the number of channels (or neurons), as each is mapped to a single L-LUT. Theoretically, each L-LUT occupies $\frac{Y}{3}(2^{X-4}-(-1)^{X})$ 6-LUTs as per \cite{logicnets}. For the input convolution: $Y{=}B_h,\, X{=}B_\text{in}F_\text{in}$; for hidden convolutions: $Y{=}B_h,\, X{=}B_hF_\text{conv}$; for pooling: $Y{=}B_h,\, X{=}B_hF_\text{fc}$; and for the final FC layer: $Y{=}B_\text{out},\, X{=}B_hF_\text{fc}$.}
\label{tab:architectures}
\resizebox{\textwidth}{!}{%
\begin{tabular}{lcccccc}
\hline
\textbf{Model} & \textbf{Bitwidth} & \textbf{Fan-in} & \textbf{Conv layers: unrolling factor$\times$(channels per layer)} & \textbf{Pool FC} & \textbf{FC} \\
\textbf{(Dataset)} & \textbf{$B_\text{in}$/$B_\text{h}$/$B_\text{out}$}  & \textbf{$F_\text{in}$/$F_\text{conv}$/$F_\text{fc}$} &\textbf{ unrolling factor $\times$ L-LUTs per layer} & \textbf{L-LUTs} & \textbf{L-LUTs}\\
\hline
CNN-S       & 4/2/3 & 2/4/4 &
  $256\times(14+56+14);\; 64\times(112+28);\; 16\times(224+56)$ &
  2{,}240 & 2{,}240\\
CNN-M      & 4/2/3 & 2/4/4 &
  $256\times(28+224+56+14);\; 64\times(448+112+28);\; 16\times(896+224+56)$ &
  2{,}240 & 2{,}240 \\
CNN-L     & 4/2/3 & 2/4/5 &
  $256\times(50+400+100+25);\; 64\times(800+200+50);\; 16\times(1600+400+100)$ &
  4{,}000 & (4{,}000)$\times$2 \\
DS-CNN-A         & 4/2/4 & 2/4/4 &
  $80\times(32+32+64);\; 20\times(64+128);\; 10\times(128+256)$ &
  1{,}536 & 1{,}536 \\
DS-CNN-B          & 4/2/4 & 2/4/4 &
  $80\times(64+64+128);\; 20\times(128+256);\; 10\times(256+512)$ &
  3{,}072 & 3{,}072 \\
DS-CNN-C         & 3/2/3 & 2/3/3 &
  $2304\times(8+8+16);\; 576\times(16+32);\; 144\times(32+64);\; 36\times(64+128)$ &
  3{,}840 & 3{,}840 \\
DS-CNN-D         & 4/2/2 & 2/4/4 &
  $2304\times(8+8+16);\; 576\times(16+32);\; 144\times((32)\times3+64);\; 36\times((64)\times3+128)$ &
  3{,}840 & 3{,}840 \\
\hline
\end{tabular}
}
\end{table*}

\begin{table*}[t]
\footnotesize
\centering
\caption{Comparison of FPGA implementations on CIFAR-10. N/R = not reported. $^\S$No pipeline registers reported; single-cycle inference assumed. $^\dagger$Theoretical gate count without the group sum layer; FPGA utilization not reported. $^*$Results taken from out-of-context. The area delay column represents the product of LUTs and latency.}
\label{tab:fpga-comparison}

\resizebox{\textwidth}{!}{%
\begin{tabular}{lrrrrrrrrrr}
\toprule
\textbf{Work} &
\textbf{Acc. (\%)} &
\textbf{LUT (\% util)} &
\textbf{FF} &
\textbf{BRAM} &
\textbf{DSP} &
\textbf{Latency (ns)} &
\textbf{Throughput (FPS)} &
\textbf{$F_{\mathrm{max}}$ (MHz)} &
\textbf{Area$\times$Delay} &
\textbf{FPGA} \\
\midrule

LL-ViT~\cite{llvit}
  & \textbf{95.5} & 589,000 & 229,000 & 1,425 & 0
  & 5,330,000,000 & 1,083 & 200 & $3.14\times10^{15}$ & VU9P \\

H-WNN L$^*$~\cite{hwnn}
  & 88.23 & 20,720 & N/R & 235 & N/R
  & 6,840,000,000 & N/R & 100 & $1.42\times10^{13}$ & Pynq-Z1 \\

H-WNN S$^*$~\cite{hwnn}
  & 84.62 & 22,870 & N/R & 110 & N/R
  & 1,600,000,000 & N/R & 100 & $3.66\times10^{12}$ & Pynq-Z1 \\

hls4ml TinyML~\cite{finn_tinyml}
  & 83.5 & 28,544 & 49,215 & 42 & 4
  & 27,300,000 & N/R & N/R & $8.82\times10^{11}$ & Pynq-Z2 \\

FINN TinyML~\cite{finn_tinyml}
  & 84.5 & 24,502 & 34,354 & 100 & 0
  & 1,500,000 & N/R & N/R & $3.99\times10^{10}$ & Pynq-Z2 \\

fpgaconvnet~\cite{fpgaconvnet_results}
  & 86.0 & 47,000 & N/R & 93 & 211
  & 410,000 & N/R & 143 & $1.93\times10^{10}$ & Z-7020 \\

FINN~\cite{finn}
  & 80.1 & 46,253 & N/R & 186 & N/R
  & 283,000 & 21.9K & 200 & $1.31\times10^{10}$ & Z-7045 \\

DWN$^*$ ($n{=}6$)~\cite{dwn}
  & 57.42 & \textbf{16,700} & N/R & N/R & N/R
  & N/R & 250M & 250 & N/R & Z-7045 \\

LogicTreeNet-B$^*$~\cite{petersen2024convolutional}
  & 80.17 & 16M$^\dagger$ & N/R & N/R & N/R
  & 24$^\S$ & 41.6M & 41.6 & $3.84\times10^{8}$ & VU13P \\

LogicTreeNet-M$^*$~\cite{petersen2024convolutional}
  & 71.01 & 3M$^\dagger$ & N/R & N/R & N/R
  & \textbf{9}$^\S$ & 111.1M & 111.1 & $2.70\times10^{7}$ & VU13P \\

\midrule
\textbf{ConvLUT CNN-S$^*$ (2 cycles)}
  & 70.78 & 290,715 (25\%) & \textbf{9,061} & \textbf{0} & \textbf{0}
  & \textbf{9} & 216M & 216 & $\mathbf{2.62\times10^{6}}$ & VU9P \\

\textbf{ConvLUT CNN-S$^*$ (10 cycles)}
  & 70.78 & 308,895 (26\%) & 90,191 & \textbf{0} & \textbf{0}
  & 30 & \textbf{337M} & \textbf{337} & $9.27\times10^{6}$ & VU9P \\

\textbf{ConvLUT CNN-M$^*$ (2 cycles)}
  & 75.86 & 1,015,299 (40\%) & 16,749 & \textbf{0} & \textbf{0}
  & 20 & 101M & 101 & $2.03\times10^{7}$ & VU440 \\

\textbf{ConvLUT CNN-M$^*$ (13 cycles)}
  & 75.86 & 1,157,752 (46\%) & 303,764 & \textbf{0} & \textbf{0}
  & 78 & 167M & 167 & $9.03\times10^{7}$ & VU440 \\

\textbf{ConvLUT CNN-L$^*$ (2 cycles)}
  & 80.08 & 2,130,538 (84\%) & 28,176 & \textbf{0} & \textbf{0}
  & 32 & 62M & 62 & $6.82\times10^{7}$ & VU440 \\

\textbf{ConvLUT CNN-L$^*$ (14 cycles)}
  & 80.08 & 2,454,784 (97\%) & 539,472 & \textbf{0} & \textbf{0}
  & 182 & 77M & 77 & $4.47\times10^{8}$ & VU440 \\

\bottomrule
\end{tabular}%
}
\label{tab:cifar}
\end{table*}

\begin{table*}[t]
\footnotesize
\centering
\caption{Comparison of FPGA implementations on keyword spotting (KWS) using the Speech Commands dataset. N/R = not reported. $^*$Results taken from out-of-context.}
\label{tab:kws-comparison}

\resizebox{\textwidth}{!}{%
\begin{tabular}{lrrrrrrrrrr}
\toprule
\textbf{Work} &
\textbf{Acc. (\%)} &
\textbf{LUT (\% util)} &
\textbf{FF} &
\textbf{BRAM} &
\textbf{DSP} &
\textbf{Latency (ns)} &
\textbf{Throughput (FPS)} &
\textbf{$F_{\mathrm{max}}$ (MHz)} &
\textbf{Area$\times$Delay} &
\textbf{FPGA} \\
\midrule

H-WNN L$^*$~\cite{hwnn}
  & 81.52 & 22,870 & N/R & 110 & N/R 
  & 1,600,000,000 & N/R & 100 & $3.66\times10^{13}$ & Pynq-Z1 \\

H-WNN S$^*$~\cite{hwnn}
  & 78.22 & 17,740 & N/R & 42 & N/R
  & 640,000,000 & N/R & 100 & $1.14\times10^{13}$ & Pynq-Z1 \\
  
fpgaconvnet~\cite{fpgaconvnet_results}
  & 91.1 & 37,000 & N/R & 97 & 188 
  & 320,000 & N/R & 143 & $1.18\times10^{10}$ & Z-7020 \\
  
FINN TinyML~\cite{finn_tinyml}
  & 82.5 & 33,732 & 34,405 & 37 & 0 
  & 17,000 & N/R & N/R & $5.91\times10^{8}$ & Pynq-Z2 \\

DWN$^*$ ($n{=}6$)~\cite{dwn}
  & 71.52 & \textbf{4,800} & N/R & N/R & N/R 
  & N/R & 250M & 250 & N/R & Z-7045 \\

\midrule

\textbf{ConvLUT DS-CNN-A$^*$ (2 cycles)}
  & 88.99 & 144,711 (12\%) & \textbf{9,615} & \textbf{0}
  & \textbf{0} & \textbf{9}
  & 208M & 208 & $\mathbf{1.30\times10^{6}}$ & VU9P \\

\textbf{ConvLUT DS-CNN-A$^*$ (10 cycles)}
  & 88.99 & 146,886 (12\%) & 50,745 & \textbf{0}
  & \textbf{0} & 21
  & \textbf{481M} & \textbf{481} & $3.08\times10^{6}$ & VU9P \\

\textbf{ConvLUT DS-CNN-B$^*$ (2 cycles)}
  & \textbf{91.53} & 286,306 (24\%) & 15,554 & \textbf{0}
  & \textbf{0} & 12
  & 173M & 173 & $3.44\times10^{6}$ & VU9P \\

\textbf{ConvLUT DS-CNN-B$^*$ (10 cycles)}
  & \textbf{91.53} & 286,611 (24\%) & 92,050 & \textbf{0}
  & \textbf{0} & 30
  & 338M & 338 & $8.60\times10^{6}$ & VU9P \\
\bottomrule
\end{tabular}%
}
\label{tab:kws}
\end{table*}

\begin{table*}[t]
\footnotesize
\centering
\caption{Comparison of FPGA-based inference implementations on visual wakeword (VWW). N/R = not reported.}
\label{tab:vww-comparison}

\resizebox{\textwidth}{!}{%
\begin{tabular}{lrrrrrrrrrr}
\toprule
\textbf{Work} &
\textbf{Acc. (\%)} &
\textbf{LUT (\% util)} &
\textbf{FF} &
\textbf{BRAM} &
\textbf{DSP} &
\textbf{Latency (ns)} &
\textbf{Throughput (FPS)} &
\textbf{$F_{\mathrm{max}}$ (MHz)} &
\textbf{Area$\times$Delay} &
\textbf{FPGA} \\
\midrule

fpgaconvnet~\cite{fpgaconvnet_results}
  & \textbf{84.8} & \textbf{34,000} & \textbf{123} & N/R & 189 
  & 9,490,000 & N/R & 111 & $3.23\times10^{11}$ & Z-7020 \\

\midrule

\textbf{ConvLUT DS-CNN-C$^*$ (2 cycles)}
  & 76.70 & 216,093 (18\%) & 35,059 & \textbf{0}
  & \textbf{0} & \textbf{7}
  & 302M & 302 & $\mathbf{1.43\times10^{6}}$ & VU9P \\
\textbf{ConvLUT DS-CNN-C$^*$ (12 cycles)}
  & 76.70 & 388,163 (33\%) & 275,134 & \textbf{0}
  & \textbf{0} & 27
  & \textbf{450M} & \textbf{450} & $1.05\times10^{7}$ & VU9P \\
\textbf{ConvLUT DS-CNN-D$^*$ (3 cycles)}
  & 80.01 & 927,118 (37\%)& 37,236 & \textbf{0}
  & \textbf{0}
  & 23 & 129M & 129 & $2.13\times10^{7}$ & VU440 \\
\textbf{ConvLUT DS-CNN-D$^*$ (16 cycles)}
  & 80.01 & 1,107,470 (44\%) & 333,359 & \textbf{0}
  & \textbf{0} & 63
  & 256M & 256 & $6.98\times10^{7}$ & VU440 \\
\bottomrule
\end{tabular}%
}
\label{tab:vww}
\end{table*}

\subsection{Effect of Iterative Pruning Retraining Budget}

\begin{figure*}[tb]
\centerline{\includegraphics[width=0.65\linewidth]{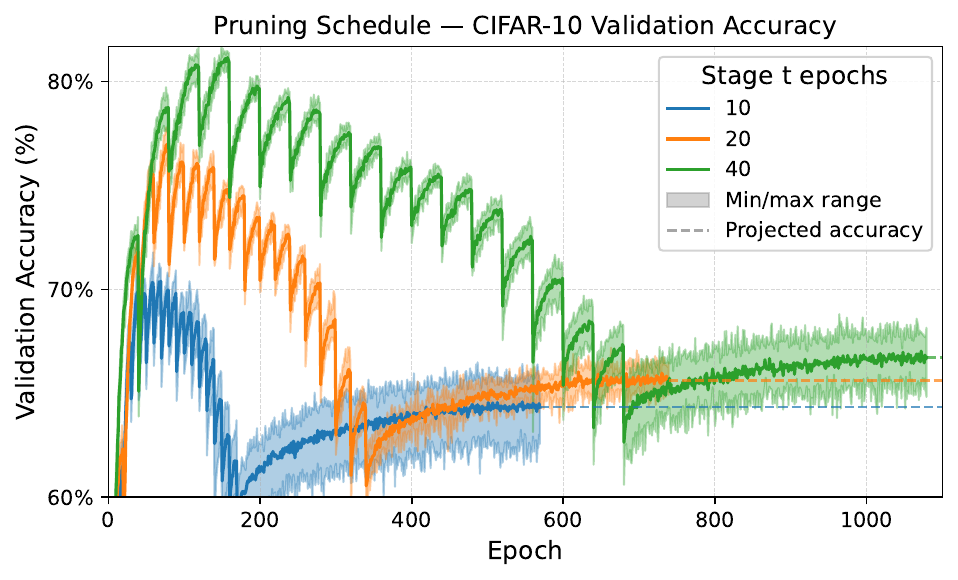}}
\caption{Effect of the per-stage retraining budget $E_t$ on CIFAR-10 validation accuracy during iterative pruning. Each sawtooth corresponds to one pruning stage: accuracy drops sharply as the fan-in is tightened from $F_t$ to $F_{t+1}$, then recovers during the $E_t$-epoch retraining window. The dashed lines indicate the projected final accuracy for each schedule.}
\label{fig:pruning_ablation}
\end{figure*}

Fig.~\ref{fig:pruning_ablation} shows the CIFAR-10 validation accuracy throughout the iterative pruning schedule for three per-stage retraining budgets: $E_t \in \{10, 20, 40\}$ epochs. The characteristic sawtooth pattern confirms the expected behavior of the algorithm: each pruning step causes an immediate accuracy drop as connectivity is tightened, followed by recovery during the subsequent retraining window. Longer retraining budgets ($E_t = 40$, green) allow greater recovery at each stage, yielding higher final accuracy after the concluding fine-tune phase. Shorter budgets ($E_t = 10$, blue; $E_t = 20$, orange) accumulate unrecovered accuracy loss across stages, converging to a lower final value. The overall 
trend is clear: the larger the per-stage budget, the higher the final accuracy as longer retraining allows greater recovery before the next connectivity reduction. However, using $E_t = 40$ slows down training in practice. To speed it up, we adopt a non-uniform schedule in our evaluation results: 40 epochs for the first five stages where fan-in reductions are largest, 10 for intermediate stages, and 20 for the final five stages before the concluding fine-tune.

\subsection{LUT-based vs Classical Pooling}
 
Table~\ref{tab:pooling_comparison} compares the proposed LUT pooling layer with classical max and average pooling. The size column reports the equivalent LUT representation, highlighting the hardware incompatibility: max and average pooling induce fan-ins of 32, 20, and 72 inputs, resulting in theoretical L-LUT sizes that are infeasible on any FPGA; hence, they could not be synthesised.

In contrast, the proposed LUT pooling layer respects the hardware fan-in budget, reducing table size by several orders of magnitude. Despite this constraint, expressivity is preserved through a learned sparse aggregation function and channel expansion ($K > 1$). As a result, LUT pooling outperforms both classical alternatives on CIFAR-10, VWW, and KWS, confirming this construction's expressiveness while satisfying the hardware fan-in constraints.

\begin{table}[h]
\centering
\footnotesize
\caption{Comparison of pooling strategies for each dataset using toy model architectures. The L-LUT size is calculated based on the fan-in and bitwidth of each neuron. The number of L-LUTs is equal to the number of neurons in the pooling layer. Size is expressed as $2^{FB} \times$ number of L-LUTs.}
\label{tab:pooling_comparison}
\begin{tabular}{ccccc}
\toprule
\textbf{Pooling} & \textbf{Metric} & \textbf{CIFAR-10} & \textbf{KWS} & \textbf{VWW} \\
\midrule

\multirow{2}{*}{Max}
& Val Acc (\%) & 55.78 & 77.77 & 74.50 \\
\cmidrule(lr){2-5}
& Size & $2^{32} \times 56$ & $2^{20} \times 256$ & $2^{72} \times 128$ \\
\midrule

\multirow{2}{*}{Avg}
& Val Acc (\%) & 60.29 & 82.21 & 74.91 \\
\cmidrule(lr){2-5}
& Size & $2^{32} \times 56$ & $2^{20} \times 256$ & $2^{72} \times 128$ \\
\midrule

\textbf{LUT}
& Val Acc (\%) & \textbf{70.91} & \textbf{83.14} &  \textbf{77.47} \\
\cmidrule(lr){2-5}
\textbf{(Ours)} & Size & $\mathbf{2^{8} \times 2240}$ & $\mathbf{2^{8} \times 1536 }$ & $\mathbf{2^{6} \times 3840}$ \\

\bottomrule
\end{tabular}
\end{table}

\section{Conclusion}
We presented ConvLUT, a family of LUT-mappable convolutional architectures that scale LUT-based neural networks beyond MLP-style models to structured vision and audio tasks. ConvLUT introduces several architectural innovations for convolutions and pooling. Coupled with a hardware-aware iterative pruning strategy, ConvLUT achieves ultra-low latency CNN implementations under strict fan-in constraints.
Across CIFAR-10, Keyword Spotting, and Visual Wake Words, ConvLUT achieves competitive accuracy while delivering latencies down to $9$\,ns and the lowest area-delay products. These results demonstrate that convolutional architectures can be co-designed with FPGA LUT primitives to achieve both scalability and extreme efficiency. This performance is achieved through fully unrolled implementations, representing the low-latency end of the area-latency design space. Alternative trade-offs can be explored by reducing the unrolling factor through rolling.

\section*{Acknowledgements}
We thank BoolSi for providing Marta Andronic with an internship opportunity that helped lay the groundwork for the initial proof of concept of this project.

\bibliographystyle{ACM-Reference-Format}
\bibliography{refs}

\end{document}